\pdfoutput=1
\documentclass[10pt, a4paper]{deepmind}

\usepackage[authoryear, sort&compress, round]{natbib} 

\usepackage[utf8]{inputenc}
\usepackage[hypcap=false]{subcaption}
\usepackage{graphicx}
\usepackage{wrapfig}
\usepackage{listings}
\usepackage{multirow}
\usepackage{amsmath,amsfonts,bm}
\usepackage{xspace}

\usepackage{calc}

\renewcommand{\today}{}
\usepackage{color}
\definecolor{codegreen}{rgb}{0,0.5,0}
\definecolor{codeblue}{rgb}{0,0,0.9}
\definecolor{codegray}{rgb}{0.5,0.5,0.5}
\definecolor{codepurple}{rgb}{0.58,0,0.82}
\definecolor{backcolour}{rgb}{0.95,0.95,0.92}
\definecolor{backcolour2}{rgb}{0.9,0.9,0.9}
\definecolor{codered}{rgb}{0.5,0,0}
\definecolor{textcodered}{rgb}{0.4,0,0}
\definecolor{palegray}{rgb}{0.98,0.98,0.99}
\lstdefinestyle{mystyle}{
    backgroundcolor=\color{backcolour},   
    commentstyle=\color{codered},
    keywordstyle=\color{codeblue},
    numberstyle=\tiny\color{codegray},
    stringstyle=\color{codegreen},
    breakatwhitespace=false,         
    breaklines=true,                 
    captionpos=b,                    
    keepspaces=true,                 
    numbersep=5pt,                  
    showspaces=false,                
    showstringspaces=false,
    showtabs=false,                  
    tabsize=2,
    basicstyle=\ttfamily\footnotesize,
}

\newcommand{\E}{\mathbb{E}}
\newcommand{\mytimes}{\medmuskip=0mu\times}

\newcommand{\tictactoe}{tic-tac-toe\xspace}
\newcommand{\go}{Go\xspace}
\newcommand{\chess}{chess\xspace}
\newcommand{\shogi}{shogi\xspace}
\newcommand{\sokoban}{Sokoban\xspace}

\newcommand{\jaco}{\textsf{Jaco}\xspace}
\newcommand{\gnugo}{\textsf{GNU Go}\xspace}
\newcommand{\mujoban}{\textsf{MuJoBan}\xspace}
\newcommand{\mujogo}{\textsf{MuJoGo}\xspace}
\newcommand{\mujoxo}{\textsf{MuJoXO}\xspace}
\newcommand{\deepmind}{\textsf{DeepMind}\xspace}

\newcommand{\repo}{deepmind-research/tree/master/physics_planning_games}

\usepackage{hyperref}
\hypersetup{colorlinks=true, allcolors=blue}

\correspondingauthor{\{mmirza, drewjaegle\}@google.com}

\author[1,*]{Mehdi Mirza}
\author[1,*]{Andrew Jaegle}
\author[3,*]{Jonathan J. Hunt}
\author[1,*]{Arthur Guez}
\author[1]{Saran Tunyasuvunakool}
\author[1]{Alistair Muldal}
\author[1]{Théophane Weber}
\author[2,3]{Peter Karkus}
\author[1]{Sébastien Racanière}
\author[1]{Lars Buesing}
\author[1]{Timothy Lillicrap}
\author[1]{Nicolas Heess}

\affil[*]{Equal contributions}
\affil[1]{\deepmind}
\affil[2]{National University of Singapore}
\affil[3]{Work done at \deepmind}

\title{Physically Embedded Planning Problems: New Challenges for Reinforcement Learning}

\begin{document}

\begin{abstract}
Recent work in deep reinforcement learning (RL) has produced algorithms capable of mastering challenging games such as \go, \chess, or \shogi. In these works the RL agent directly observes the natural state of the game and controls that state directly with its actions. However, when humans play such games, they do not just reason about the moves but also interact with their physical environment. They understand the state of the game by looking at the physical board in front of them and modify it
by manipulating pieces using touch and fine-grained motor control. Mastering complicated physical systems with abstract goals is a central challenge for artificial intelligence, but it remains out of reach for existing RL algorithms. To encourage progress towards this goal we introduce a set of physically embedded planning problems and make them \href{https://github.com/deepmind/\repo}{publicly available}. We embed challenging symbolic tasks (\sokoban, \tictactoe, and \go) in a physics engine to produce a set of tasks that require perception, reasoning, and motor control over long time horizons. Although existing RL algorithms can tackle the symbolic versions of these tasks, we find that they struggle to master even the simplest of their physically embedded counterparts. As a first step towards characterizing the space of solution to these tasks, we introduce a strong baseline that uses a pre-trained expert game player to provide hints in the abstract space to an RL agent’s policy while training it on the full sensorimotor control task. The resulting agent solves many of the tasks, underlining the need for methods that bridge the gap between abstract planning and embodied control. See illustrating \href{https://youtu.be/RwHiHlym_1k}{video}.
\end{abstract}

\maketitle

\section{Introduction}

Early work in artificial intelligence (AI) often focused on symbolic reasoning and ignored the challenges of physical control and grounding, which were sometimes discounted as less central to the problem of intelligence~\citep{brooks90elephants}. Since then, many approaches to physical control problems have been developed in different disciplines, including  optimal control, robotics, and reinforcement learning, striking different trade-offs between learning and human engineering.

In recent years deep reinforcement learning (RL) has shown promising results on a number of problems that exemplify different aspects of artificial and embodied intelligence.
For instance, results on difficult board games such as \chess, \go \citep{silver2018general}, and \sokoban \citep{racaniere2017imagination} have demonstrated that RL can solve tasks which, due to their combinatorial structure and the irreversiblity of decisions, are often considered as \emph{reasoning} or \emph{planning} domains. In contrast, results such as those of \citep{levine2016endtoend, heess2017emergence, openai2018learning, hwangbo2019learning, merel2020reusable} demonstrate that RL algorithms can be successfully applied to problems that involve sophisticated \emph{low-level control} of physically embodied systems, whether in manipulation or locomotion tasks that require the short-term coordination of high-dimensional bodies. RL methods have also been able to tackle domains where \emph{perception} is a challenge, due to partial observability or a complex sensory stream \citep{beattie2016dmlab, wayne2018unsupervised, gregor2019shaping, song2020vmpo, lepaine2020making}.

However, many real world problems, or ethologically relevant tasks, require a combination of high-level reasoning, low-level motor control, and perception at very different temporal scales.
This is well recognized in the robotics literature where long-horizon sequential reasoning has long been studied \citep[e.g.][]{kambhampati1991combining}. Through the use of hand-crafted and structural methods (which often rely on the use of privileged information), this research program has led to many impressive results, including on challenging multi-object manipulation tasks \citep[e.g.,][]{dantam2018task, lagriffoul2018platform} and on games such as chess \citep{dantam2011motion}. 
In contrast, learning-based methods have so far paid relatively little attention to problems that combine all of the challenges mentioned above. Although progress has been made along each individual axes of complexity, domain agnostic RL methods are not competitive on long-horizon embodied planning problems against the well-engineered, domain-specific approaches common in the robotics literature. 

The aim of this work is to facilitate the study of such problems with learning based approaches. We provide a set of simulation domains that combine the difficulties of planning with those of low-level motor control. This is achieved by embedding existing planning tasks (\tictactoe, \go, \sokoban) into a physical setting where, in order to execute a move in the game, the agent is required to control a physical body over many simulation time-steps. 
For example, placing a single \tictactoe piece requires reaching the board with a 9 DoF (degree of freedom) arm and touching the appropriate position on the board. Although learning to play \tictactoe and executing goal-directed reaching movements with a robotic arm are individually well within the capabilities of current deep RL techniques, combining these two control problems into a single task can pose challenges. 
This effect is stronger for more difficult planning tasks (e.g. going from \tictactoe to \go) and for problems that require more sophisticated perception or low-level control (e.g., egocentric vision or controlling a physical body with many DoFs).

Our tasks complement popular motor control tasks that are currently used as benchmarks by the RL community \citep[e.g.][]{brockman2016gym,tassa2018control}. These tasks often focus on the short-term motor control component in the form of locomotion or manipulation, but require only limited high-level reasoning. They tend to be reactive with a small effective decision horizon, and the solution can often be memorized once the right behavior is found. In contrast, our domains force the agent to respond to a combinatorial number of situations, by achieving a sequence of well-defined abstract subgoals, which need to be chosen while taking into account their long-term, and possibly irreversible, consequences. The combination of motor control and high-level reasoning demands an agent that can effectively learn to act at multiple time-scales. Our benchmark is designed with the goal of providing a fertile experimental setting for the development of RL methods for physically-grounded sequential reasoning and control, rather than to develop methods that can be directly ported to physical robots. In this respect, our benchmark is complementary to others that strike a different balance between physical and abstract reasoning complexity by more realistically simulating laboratory robotic platforms \citep{dantam2018task, lagriffoul2018platform}.

Our tasks embed abstract planning problems into a physical environment with the need for motor control. This distinguishes them from tasks in which the physical environment and the reasoning problem are more intimately coupled \citep[e.g.][]{lee2019ikea, fazeli2019jenga}. Although this design choice limits the scope of the tasks we propose, it also has several key advantages: 1) the properties and requirements of the underlying reasoning task are relatively well understood,
2) the tasks possess natural abstractions which we can exploit, 3) we know RL methods that can tackle both the high-level and low-level problems in isolation, 4) we can vary the difficulty of the underlying reasoning task while keeping the control difficulty roughly constant, and conversely 5) we can vary the difficulty of the sensorimotor RL task (and how tightly it corresponds to the abstract task) while keeping the difficulty of the underlying reasoning task constant.

As a reference, and to demonstrate the general feasibility of our tasks, we include a preliminary investigation of training RL agents on our domains in Section \ref{sec:Experiments}. We study vanilla model-free agents as well as agents that have access to the output of a pre-specified abstract planner \citep[similar to the approach used by ][]{akkaya2019solving}. This study demonstrates that these domains remain challenging even when considerable privileged information about the structure of the problem is available. The insights gained from these experiments may inform the search for methods that can solve such domains efficiently.

\section{Environments}
\label{sec:Environments}

We introduce a set of three discrete planning problems embedded in continuous physical domains, illustrated in Figure \ref{fig:mujoxogoban}. These environments are open-sourced \href{https://github.com/deepmind/\repo}{here} and code snippets are available in Appendix Section~\ref{sec:appendix_code}.

\begin{figure}[htb]
    \begin{subfigure}[b]{0.5\textwidth}
    \centering
      \includegraphics[width=0.8\linewidth]{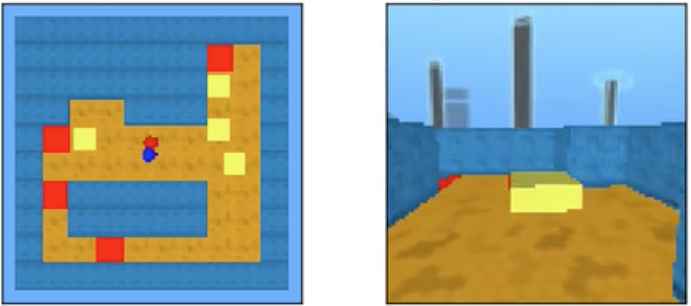}
      \caption{}
    \end{subfigure}
    \begin{subfigure}[b]{0.5\textwidth}
    \centering
    \includegraphics[width=0.8\linewidth]{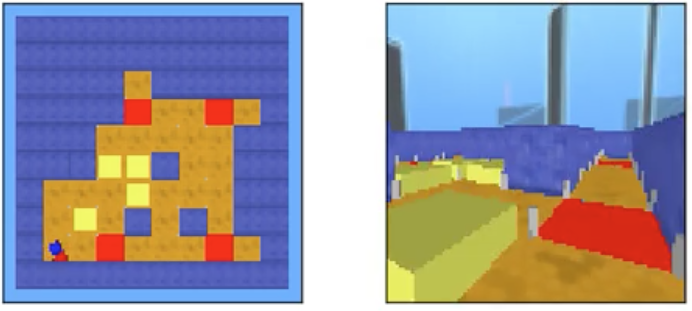}
      \caption{}
    \end{subfigure}
    \caption{\textbf{(Left pair)} \mujoban without grid pegs, example of the top-down view and the corresponding agent's egocentric view next to it. \textbf{(Right pair)}: Same camera views for \mujoban with grid pegs.}
    \label{fig:mujoban}
\end{figure}

\subsection{\mujoban}

\mujoban is a single player puzzle-solving game embedded in the MuJoCo simulation environment~\citep{todorov2012mujoco}. 
The puzzle is based on the 2D game of \sokoban (see \href{https://en.wikipedia.org/wiki/Sokoban}{here}), where an agent situated on a grid has to push boxes onto target locations. Only one box can be pushed at a time, and boxes can only be pushed but not pulled. This rule leads to many actions being irreversible (e.g.\ pushing a box into a corner), so an agent must choose its actions carefully and deliberately to solve the game. This is why \sokoban has proven popular as an RL benchmark for studying planning-type methods~\citep{junghanns1997sokoban, quack2015simultaneously, racaniere2017imagination, guez2019investigation}. 

In \mujoban, an embodied agent is situated in the 3D maze equivalent of a \sokoban level. The visual appearance of the environment (such as color of walls) is varied from episode to episode. To accomplish the task, the agent must use observations of the environment to push boxes onto targets with its physical body. The level of difficulty of the task can be varied by changing the size and difficulty of the maze, the agent's body, and the observations available to the agent.

In the base version of \mujoban, the agent has a 2-DoF ball body that is able to move boxes by rolling into them (we use the `JumpingBallWithHead' body described in \cite{tassa2020dmcontrol} but with the jump action disabled). Boxes can move horizontally along the floor (including diagonally), but are constrained to not rotate. The agent can move boxes by pushing them with its body. Observations available to the agent include standard proprioceptive observations (see section \ref{sec:other_experimental_details} for details), an egocentric camera (from the agent's perspective), and a top-down camera that shows the entire level (see Figure \ref{fig:mujoban}). To allow the agent to infer its orientation from the egocentric camera, the maze is surrounded by a fixed backdrop with a distinctive skyline pattern.
In order to infer its orientation from the top-down camera view, the agent is topped with two differently colored `ears' (a blue left ear and a red right ear), whose relative position indicates the agent's current orientation. If the agent solves the puzzle by moving each box onto a pad, the agent receives a reward of 10 and the episode terminates. The agent also receives a shaping reward of 1 for pushing a box onto a pad or of -1 for pushing a box off of a pad. The agent receives 0 reward at all other transitions. The episode terminates if the level is unsolved after 45 seconds (900 control steps). While many easy levels can be solved in fewer than 10 seconds, more difficulty levels sometimes require close to the full 45 seconds to be completed.

Because the game runs in a physics engine, it is difficult or impossible to implement the rules of the abstract \sokoban perfectly. In this sense, \mujoban relaxes the rules of \sokoban, and hence agents may ‘cheat` by using moves that are impossible in \sokoban to solve the task in \mujoban. For example, in some situations, an agent can `pull' a stuck box out of a corner by rubbing against the box's side and nudging it away from the wall with the resulting friction. In the base version of the game, we find that agents often solve the task by simultaneously pushing more than one box, by using diagonal movements to move between boxes, or by pushing boxes past each other while allowing them to partially occupy the same grid position in state space (some examples are shown in the video \href{https://youtu.be/RwHiHlym_1k?t=60}{here}).

To mitigate these problems, we also introduce a variant of the game in which physical pegs are inserted on the floor at every grid intersection point. Boxes cannot go through pegs, which makes diagonal motion effectively impossible. This forces the agent to carefully move boxes from one grid position to another in parallel to the axes of the grid. This game variant also makes it harder for the agent to push more than one box at once, and in practice we observed that agents solve this task using a strategy that more closely resembles a legal \sokoban strategy. Because of this, this game variant is also harder to solve.

The difficulty of \mujoban can be varied in several other ways. First, the agent's body can be easily swapped with any other available locomoting agent,\footnote{Other bodies (`walkers') are available in the \deepmind control suite's \href{https://github.com/deepmind/dm_control/tree/master/dm_control/locomotion}{locomation task library}. We provide example code in the appendix showing how to configure \mujoban to use a different body.} which can provide alternative, harder variants of \mujoban, or tasks  for transfer learning.
Finally, while we test agents with access to all observations, variants could be made by restricting access to only a subset. In particular, not providing the top-down camera makes the problem more strongly partially observed, since the entire configuration of the maze (and the target configuration of the maze) cannot be inferred from a single first-person camera view. Level difficulty can be controlled by varying the size of the grid and the number of boxes per level. We procedurally generate levels as in~\citep{racaniere2017imagination, guez2019investigation}. These reference levels fit within a $10\times10$ grid, contain 4 boxes, and are of varying difficulty. Training with smaller, easier, levels is also supported, for example to create a curriculum.

\subsection{\mujoxo}

\begin{wrapfigure}{r}{4cm}
\centering
    \includegraphics[width=0.9\linewidth]{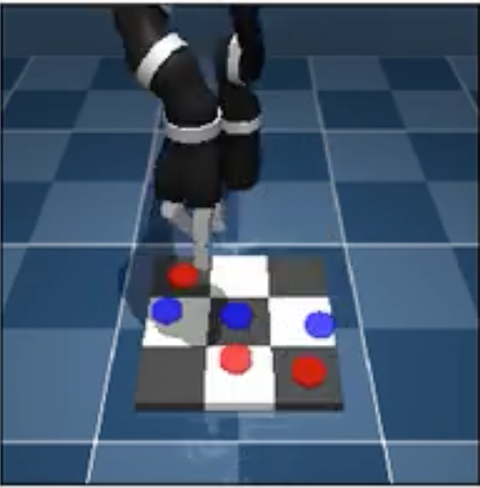}
    \caption{Example of the agent's camera view in \mujoxo.}
    \label{fig:mujoxo}
\end{wrapfigure}

\mujoxo embeds the classic game of \tictactoe in a simulated physical environment. The agent plays by controlling a robotic arm to touch board positions on a \tictactoe board presented in front of the arm. Invisible touch pads are present around each location where a piece can validly be positioned on the board. The arm is based on a 9-DoF \jaco arm \citep{campeau2017jaco}, controlled at 20Hz. When the arm makes contact with a touch pad at a position corresponding to a legal move, the move is registered and a piece (of the agent player's color) appears at the pad's physical location. Immediately following such an event, an opponent move is decided by an abstract planner and a piece is placed near the chosen location. The position of the piece on the game board is perturbed with random noise so that the pieces do not align perfectly on the grid (just as we would expect in a real-world realization of the game). After every move, and at the start of every episode, the arm is placed in a random configuration above the board.\footnote{This simplifies the problem since it avoids the need to lift or untangle the arm after a move. This reset can easily be disabled.} 
To allow for different levels of difficulty, the epsilon-greedy opponent plays a random legal move with a configurable probability $\epsilon$\ at each step, and an optimal move otherwise. 

The inputs available to the agent are composed of proprioceptive sensory data (joint angles, velocities, and torques), the global coordinate of the end effector position, the abstract state of the board, as well as a camera facing the arm and the board which provides a RGB input (84x84 pixels by default). Although we provide all these inputs to the agent, some might not be strictly necessary and may simplify the task. A subset of them can be selected for increased realism and difficulty, but we found this version to be sufficiently challenging (see Section~\ref{sec:results}).
Rewards are 0 for the duration of the episode, and a reward of 1 is given at the end of the game in case of a win or 0.5 in case of a draw.
Episodes terminate after the game is decided or after 30 seconds. Games played with trained agents finish in $\sim$5 seconds, or $\sim$100 agent steps.

\subsection{\mujogo}

\begin{figure}[h]
    \begin{subfigure}[b]{0.5\textwidth}
    \centering
      \includegraphics[width=0.4\linewidth]{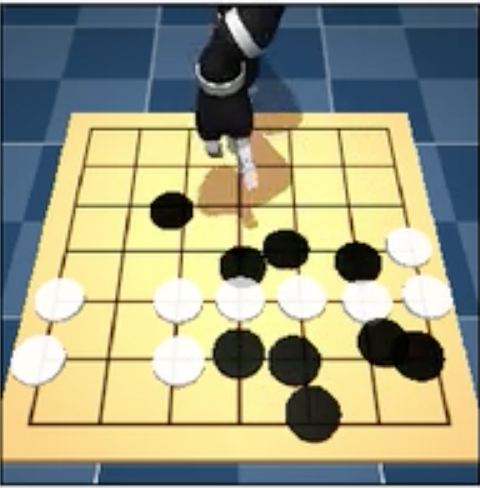}
      \caption{}
    \end{subfigure}
    \begin{subfigure}[b]{0.5\textwidth}
    \centering
    \includegraphics[width=0.65\linewidth]{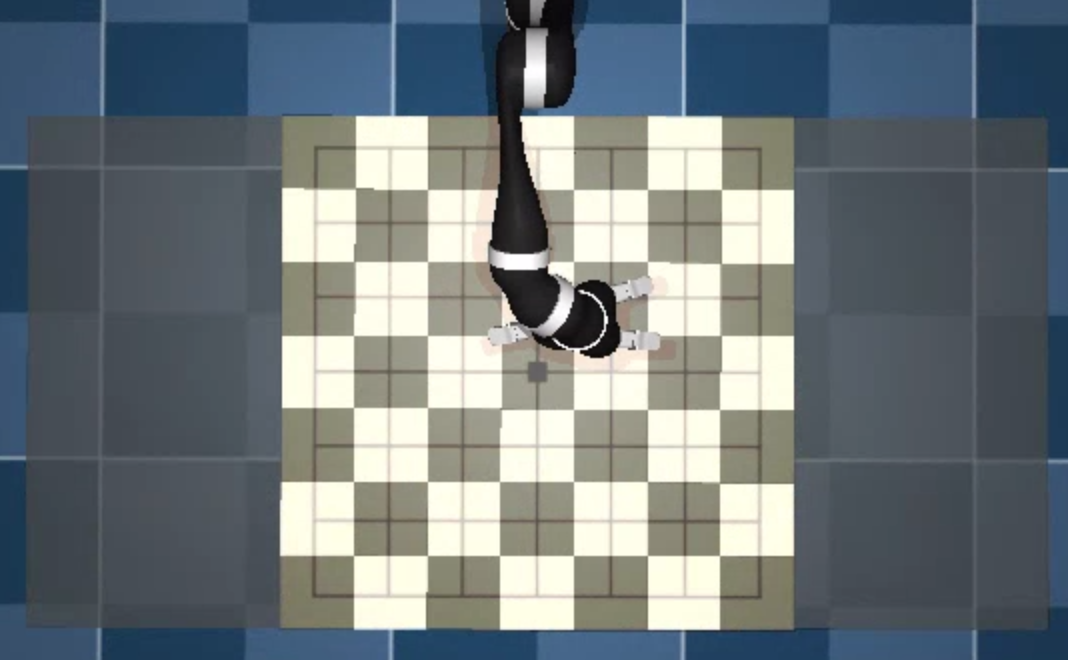}
      \caption{}
    \end{subfigure}
    \caption{\textbf{Left}: example of the agent's camera view in \mujogo. \textbf{Right}: Top-down view of the board with all touch pads visualized. The `pass' touch pads are shown as dark gray regions to the left and right of the board, and the touch pads for each intersection are shown as a checkerboard.}
    \label{fig:mujogo}
\end{figure}

\mujogo is set up similarly to \tictactoe, with the difference that the board size is a 7x7 \go board\footnote{Here, we focus on the 7x7 board size, as we found it challenging for current algorithms. However, the code for this environment allows for larger board sizes, so the difficulty of this task can be easily scaled up in the future.} and touches are registered around grid intersections. In addition, large touch pads on the left and right sides of the board are used to register pass moves. To encourage more consistent reaching movements across long \go games, only the end effectors can register a move (rather than any part of the arm, as in \mujoxo). 
Episodes terminate after 45 seconds during training, but we allow for up 60 seconds per episode when evaluating agents. Timed out games are considered losses. An efficient agent can complete a game of \mujoxo in roughly 10 moves (about 0.5 seconds) or a game of \mujogo in roughly 50 moves (about 2.5 seconds).
Games are scored according to Tromp-Taylor rules \citep{tromptaylor} with a komi of 5.5. 
The opponent's move selection is decided using the \gnugo program, a traditional \go program \citep{GNUGo2009}. Opponent strength is configured both with the \gnugo difficulty level $L$ and the probability $\epsilon$ of making a random legal action.\footnote{Communication with \gnugo is made through the Go Text Protocol (GTP), other \go programs supporting GTP could easily be used instead.} Captures are implemented by removing the appropriate stones from the board after each move.

To handle the underlying abstract game logic of \tictactoe and \go, we leverage the OpenSpiel framework \citep{lanctot2019openspiel}.
For our domains we chose to provide an opponent in the form of a configurable game engine rather than relying on self-play (as in \citealt{silver2017mastering}), to  make our tasks more compatible with single-agent training frameworks.
While we focus on \tictactoe and Go in this report, other board games could be straightforwardly included in this challenge suite. In their current form, both \mujoxo and \mujogo employ certain simplifications: for instance, there are no collisions between board pieces (they are allowed to overlap on the board). Furthermore, picking and placing of pieces with the robotic hand is not required. Finally, the agent always plays the first move (black stones in \go). These simplifications could be removed in future versions of the tasks. 

\section{Background and related work}

In this report, we describe a suite of domains designed to spur developments in long-horizon, embodied reasoning. While the components of this problem have been approached from many angles, little work in the RL and deep learning community has attempted to take on the problem as a whole. We believe that this is at least in part due to the paucity of suitable benchmarks. In this section, we review several recent approaches to motor control and long-horizon reasoning. This summary is not meant to be exhaustive, but rather to give a glimpse of the set of approaches that may be important for mastering domains like \mujoban, \mujoxo, and \mujogo without using privileged information.

\subsection{Challenges in motor control}
The last few years have seen significant interest in approaches for learning control of simulated or real robots and other embodied systems. This includes, in particular, a significant body of work on robotic arm manipulation at different levels of complexity \citep[e.g.,][]{popov2017dexterous,zhu2018reinforcement,riedmiller2018learning,lee2019silo}. Different publications consider control problems and solution strategies that emphasize different dimensions of the problem. These include a focus on dexterity
\citep[e.g.,][]{openai2018learning,akkaya2019solving}, but also long-horizon, multi-step control  \citep[e.g.,][]{gupta2019relay,duan2017one,xu2017neural}. Compared to the environments presented in this report, the combinatorial complexity and  horizon of most tasks considered in the literature is low. There is also a growing interest in multi-step, long-horizon control for locomotion and whole body control. Work such as \citep{heess2017emergence,peng2018deepmimic} considered agile but primarily short-horizon, reactive locomotion behavior. More recent work such as \cite{merel2018hierarchical,merel2020reusable} emphasize locomotion and whole-body control embedded in more complex visuomotor-reasoning tasks or multi-agent strategies  \citep[e.g.][]{liu2019emergent}.

Many of the problems studied in the motor control and robotics literature are challenging because of the difficulty of making conceptually simple, but high-level decisions in realistic state and action spaces, which may be continuous and high-dimensional. Consider a typical manipulation problem requiring a robot to pick up an object and place it in a box. This problem couples perception and fine-grained motor control. The underlying problem can in principle be solved with a simple procedure once the appropriate abstractions are identified (e.g. ``move over the object, pick up the object, move over the box, drop the object''). However, non-trivial sequences of motor commands may be required to execute each abstraction. The problem is difficult largely because it requires identifying the underlying structure through perception and correctly executing a procedure with motor control.

\subsection{Challenges in long-horizon reasoning}

In contrast, in problems such as \go and \chess a game is won or lost only after a long chain of decisions and actions, and the player may face a combinatorial number of game states that they have to respond to. For a plan to be successful, it typically needs to be executed with a high level of precision over intervals spanning a large number of decisions, as each decision can have a large impact on the final outcome, and the effects of a poorly made decision can be very difficult to recover from. These problems are notoriously difficult for RL algorithms because of the wide variety of situations that can be encountered, the difficulty of finding successful plays by chance, and the long delay between actions and rewards \citep{sutton2018reinforcement,arjona2019rudder,hung2019optimizing}. This difficulty is exacerbated when planning problems are embedded in physical domains, as many small control steps may be required to carry out a single high-level decision. Waiting to reinforce the rare, random actions that lead to rewards may be theoretically possible, but it is unworkable in practice \citep{sutton2018reinforcement,hung2019optimizing}. 

The robotics community has long recognized the importance of solving the joint problem of task-driven reasoning and movement/control (together known as \textit{task and motion planning}; \cite{yap1986algorithmic, latombe1991robot, gordon1991motion, kambhampati1991combining}). Researchers in robotics have studied these problems using reasoning domains that require long-horizon planning and multi-object sequential manipulation \citep[e.g.,][]{dantam2018task, lagriffoul2018platform}, and progress in this area has largely been driven by the design and engineering of domain-specific hand-crafted and structured methods \citep{simeon2004manipulation, siddharth2014combined, plaku2010sampling, kaelbling2011hierarchical, toussaint2015logic, dantam2016incremental}, which typically rely on access to privileged information. Although recent approaches to task and motion planning have incorporated methods from deep learning and RL \citep[e.g.,][]{chitnis2016guided, kim2019learning, driess2020deep, kim2019adversarial}, research on task and motion planning has largely occurred independently of developments in RL, and benchmarks from task and motion planning are typically not designed with RL methods in mind.

In the RL literature, a number of (often overlapping) lines of research consider strategies that may eventually address this challenge. For instance, research in model-based planning \citep[e.g.,][]{sutton1991dyna,silver2016mastering,silver2017mastering,silver2018general, chua2018deep, wang2020exploring, coreyes2018selfconsistent, finn2017deep} studies how models can be used to identify good actions either to update the policy or to structure interactions with the environment. The use of models may improve data efficiency or facilitate transfer and may be particularly useful in settings with combinatorial action choices and the need to make decisions that are not easily undone. Similarly, techniques developed in the hierarchical reinforcement learning literature \citep[e.g.,][]{dayan1993feudal,sutton1999between,bacon2017option,florensa2017stochastic, hausman2018learning, merel2018hierarchical, nachum2019near, haarnoja2018latent} focus on structured behavior representations that may encourage learning by making behaviors more reusable, by providing structure for the credit assignment or exploration process, or by otherwise improving the abstractions learned by RL.

One major challenge when solving long-horizon reasoning problems is the so-called ``exploration problem'', i.e. how to discover the rare strategies that lead to rewarded behaviors. While robust and directed exploration behavior can in principle arise solely by generalizing from relevant past experience, it is often difficult in practice. A range of methods have been proposed to bootstrap the learning process when training from scratch.
One common solution is to use shaping rewards \citep[e.g.,][]{popov2017dexterous}, i.e. rewards that are widely available in state space and which provide local information that guides the agent towards the solution, thereby shortening the effective horizon for credit assignment.
Another common solution is to use a  curriculum 
\cite[e.g.,][]{graves2017automated,riedmiller2018learning,florensa2017reverse}, i.e. a set or sequence of tasks of increasing difficulty chosen such that mastering the simple tasks makes it easier for the agent to learn solutions to harder tasks, thus also guiding the agent to a solution of the target task. 
A similar role can be played by demonstrations of expert behaviour, which can be used to provide examples of trajectories that lead to reward or to initialize policies. This is a common approach \citep{nair2018overcoming}, and has been used as a part of strategies for tackling \textsf{StarCraft} \citep{vinyals2019grandmaster}, \go \citep{silver2016mastering}, as well as motor control tasks \citep{ho2016generative, peng2018deepmimic, merel2020reusable}.
Shaping rewards, curricula, and expert behavior explicitly or implicitly encode knowledge about the solution to a given task. An alternative line of work focuses on methods that drive the agent to search over the state-action in an effective manner \citep[e.g.,][]{schmidhuber1991curious,thrun1992efficient}. One popular way of doing so to use intrinsic reward functions that incentivize the agent to visit different parts of the state space  \citep[e.g.,][]{gregor2016variational,hausman2018learning,florensa2017stochastic,bellemare2016unifying,strehl2008analysis}.

\subsection{Related RL benchmarks}

The growing interest in motor control problems that require a combination of advanced motor skills, perception, and longer-horizon reasoning has led to the demand for suitable benchmark tasks. Whereas initial task suites focused on short-horizon single task challenges \citep{brockman2016gym,tassa2018control}, more recently released benchmark suites \citep[e.g.][]{fan2018surreal,yu2019meta,james2019rlbench} focus on multi-task scenarios, where the same embodiment needs to solve a diverse set of challenges or is required to adapt existing skills to a new problem. Similar to our domains, the recently released \textsf{IKEA} furniture assembly benchmark \citep{lee2019ikea} emphasizes long-horizon reasoning, but does so in the context of an embodied construction task and provides a different trade-off in terms of perceptual difficulty, motor control, and the combinatorial complexity of the underlying planning problem. The \deepmind locomotion suite and soccer environment \citep{liu2019emergent, tassa2020dmcontrol} include both short-horizon, agility-focused problems as well as longer-horizon and multi-agent challenges. Complementary to this task suite, which primarily focuses on motor control and multi-agent coordination, the tasks we introduce couple perception and physical interaction with a challenging abstract structure in a single-agent setting.

The above summary only provides a glimpse of the relevant body of work in the RL literature. With the environments described in section \ref{sec:Environments} we provide a test bed that we hope will allow for systematic comparisons of existing approaches and facilitate the development of new approaches, thus accelerating research into difficult embodied reasoning problems combining high-level reasoning, low-level motor control, and perception.

\section{Experiments}
\label{sec:Experiments}

The domains introduced in the previous section require the agent to learn a robust motor policy and align this policy with a high-level strategy that solves the puzzle (in the case of \mujoban) or wins the game (in the case of \mujoxo and \mujogo). The goal of this section is to establish baselines for these domains for future comparisons, and also as a way to better grasp the challenge behind these tasks. As we will show, na\"ive end-to-end training with state-of-the art RL algorithms is largely unsuccessful, but an actor-critic architecture that incorporates expert planner information can learn good policies for each of these domains. This suggests that deep RL methods can learn appropriate motor plans when provided with shaping rewards that reflect high-level state and strategy information. 

To better understand the shortcomings of standard deep RL methods in these domains, we examine how much expert knowledge about the underlying abstract task (its state, dynamics, or solution) is needed to enable effective learning. To do so, we probe the performance of sensorimotor agents with access to three levels of information: \textbf{(i)} information about the abstract state, abstract dynamics, and the solution (the \textit{expert planner} condition), \textbf{(ii)} information about the abstract state and dynamics, but not the solution (the \textit{random planner} condition), \textbf{(iii)}, minimal or no information about the abstract state (beyond the sensory observations) and no information about the dynamics or solution (the \textit{vanilla agent} condition). We next describe the basic structure of our architecture (in section \ref{sec:agent_architecture}) before describing the details of the experimental conditions (in section \ref{sec:conditions}).

\begin{figure}[h]
  \centering
  \includegraphics[width=1.0\linewidth]{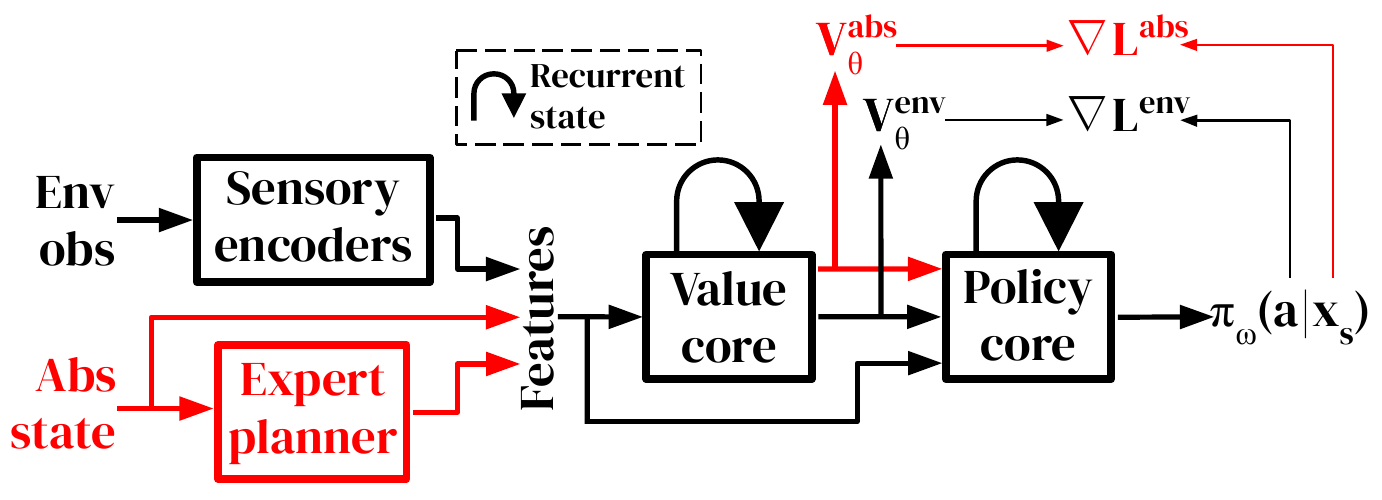}
  \caption{The agent architecture used for the `expert planner' condition. The standard components of the actor-critic policy gradient architecture are shown in \textcolor{black}{\textbf{black}}, while the components of the expert planner and the corresponding policy gradient computation are shown in \textcolor{red}{\textbf{red}}. The `vanilla agent' condition uses only the components shown in black. The `random planner' condition includes the components shown in red, but uses a random planner in place of the expert planner. Details of the architecture are given in Section \ref{sec:appendix_obs_arch} of the Appendix.} 
  \label{fig:mujoban_agent_architecture}
\end{figure}

\subsection{Agent architecture}
\label{sec:agent_architecture}
Our agent architecture is built around an actor-critic network structure similar to others used in the literature on RL for continuous control (e.g. \cite{heess2017emergence}). This architecture consists of several submodules designed for sensory encoding, value estimation, and continuous policy output. The full network architecture is trained end-to-end using the distributed \textsc{IMPALA} actor-critic algorithm with V-trace off-policy corrections \citep{espeholt2018impala}.

There are two key differences between our architecture and those used elsewhere in continuous control. The first difference is the inclusion of an expert planner module that maps ground truth abstract states to target abstract states and the game actions needed to reach them. In the context of the continuous control task the agent needs to perform, the information fed into and returned from the expert should be viewed as \emph{privileged} information. This is because the abstract game state and the action an expert would choose may be hard or impossible to infer from sensorimotor observations, but they are sufficient to solve the underlying abstract problem. We include this component to allow us to analyse the role that knowledge about the structure of the underlying state space and abstract actions plays in solving games that are grounded in a physical environment.

The second difference is the addition of an auxiliary task to follow the expert's instructions in the abstract state space. In particular, the agent receives a reward of constant magnitude when the actions executed by the policy result in the state-space transition suggested by the expert. This reward is provided if the agent follows the suggestion within a fixed time limit. In addition to providing the agent with information about the solution to the task, this reward provides the agent with information about the structure of the abstract state space and the effect of low-level actions in this state space. We structure the auxiliary task using episodes that are shorter than those of the main task and which occur throughout the main episode (each auxiliary task episode can be thought of as a sub-episode of the main episode). The target of an auxiliary task is set by the planner at the beginning of each auxiliary task episode. Auxiliary task episodes are reset when the agent reaches the target or when the time limit of the auxiliary episode is reached. Auxiliary task episodes are produced until the end of the main task episode is reached.

In early experiments, we estimated a single value that reflected the contribution of both the main task and the auxiliary task. We found that the agent struggled to learn in this setting. We believe this occurs because mixing the values for the two tasks makes it impossible to correctly bootstrap on either task and because the very different timescales of the two rewards makes it hard to balance their contributions to a single value. To encourage the agent to maximize both objectives, we estimate the value contributions of the two terms using two distinct network outputs (see Fig. \ref{fig:mujoban_agent_architecture}) and optimize the two value estimates separately. The addition of a second value estimate more cleanly separates the value associated with following the planner's suggestions from that associated with the task reward. It also allows us to handle discounting, bootstrapping, and episode termination for the two objectives separately. We found this to lead to more stable training. The agent architecture, including the two value estimates, is illustrated in Fig. \ref{fig:mujoban_agent_architecture}.

The final agent update is produced by adding the policy gradient contributions of the two objectives (each of which uses the advantage estimates from the corresponding value output):

\begin{align} 
\nabla\mathcal{L}^{env} &= \E_{x_s, a_s}\left[ \rho_s \nabla_{\omega} \log \pi_{\omega}(a_s|x_s)(r_t^{env} + \gamma^{env} v^{env}_{s+1} - V^{env}_{\theta}(x_s)) \right] \label{eq1} \\
\nabla\mathcal{L}^{abs} &= \E_{x_s, a_s}\left[ \rho_s \nabla_{\omega} \log \pi_{\omega}(a_s|x_s)(r_t^{abs}+ \gamma^{abs} v_{s+1}^{abs} - V_{\theta}^{abs}(x_s)) \right]  \label{eq2} 
\end{align}

\begin{figure}[h]
\centering
    \begin{subfigure}[b]{0.25\textwidth}
    \centering
      \includegraphics[width=0.8\linewidth]{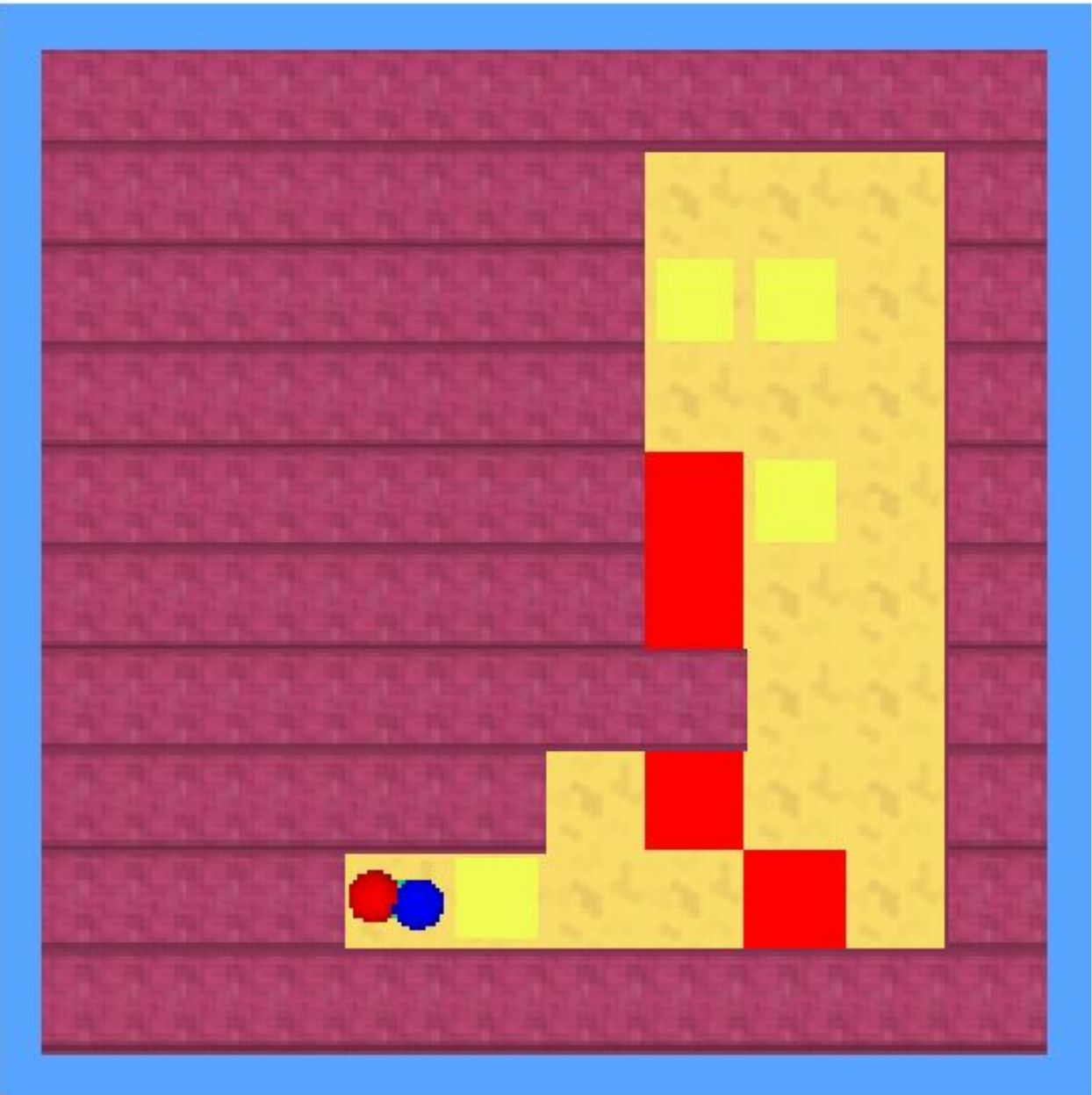}
      \caption{}
    \end{subfigure}
    \begin{subfigure}[b]{0.25\textwidth}
    \centering
    \includegraphics[width=0.8\linewidth]{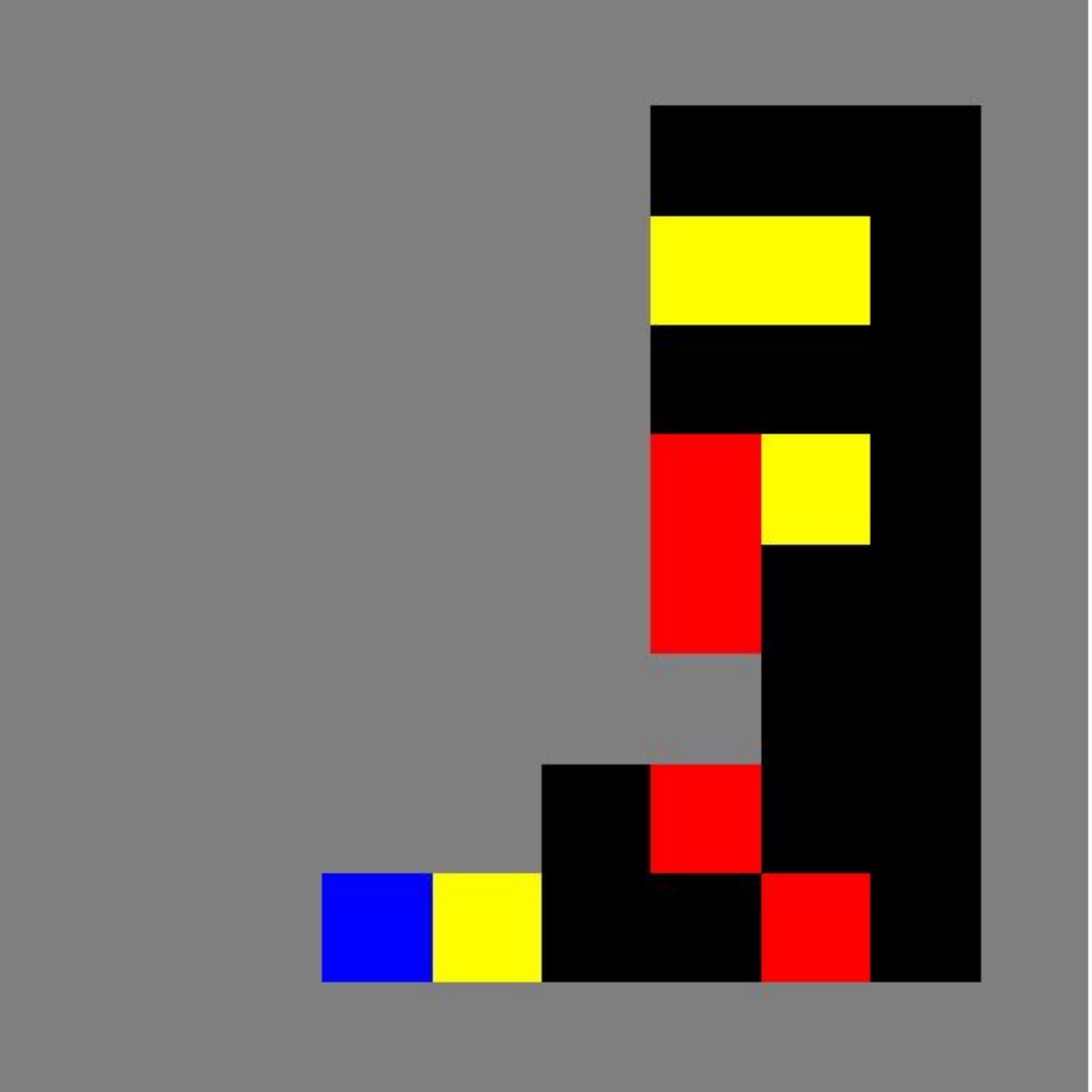}
      \caption{}
    \end{subfigure}
     \begin{subfigure}[b]{0.25\textwidth}
    \centering
    \includegraphics[width=0.8\linewidth]{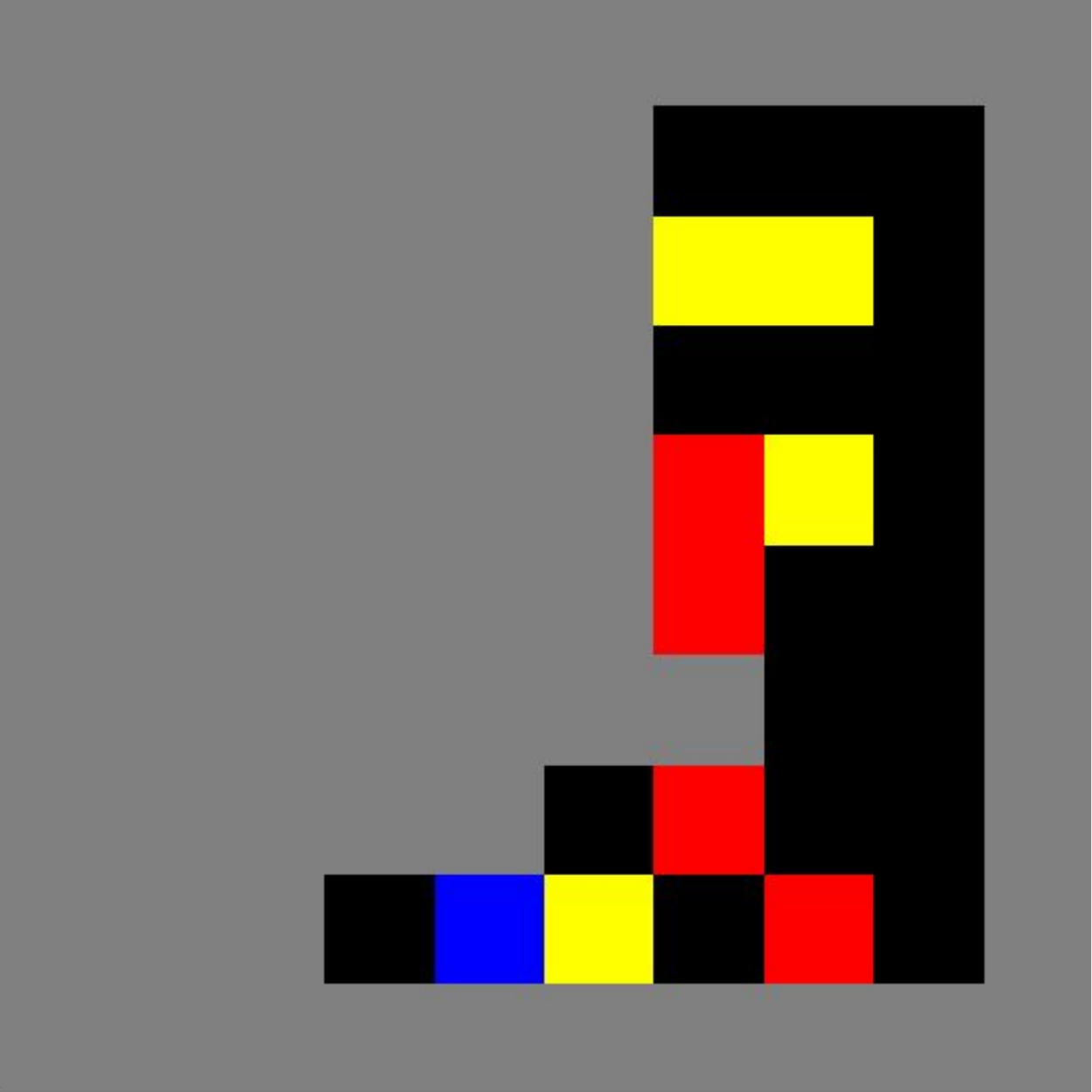}
      \caption{}
    \end{subfigure}
    \caption{Illustration of different inputs in \mujoban: (\textbf{left}) top-down camera, (\textbf{middle}) the corresponding abstract state, and (\textbf{right}) the target state provided by the expert planner. Color coding for the abstract state is as follows, \textcolor{darkgray}{grey}: walls, \textcolor{blue}{blue}: agent, \textcolor{yellow}{yellow}: boxes, \textcolor{red}{red}: box targets.}
    \label{fig:mujoban_state}
\end{figure}

Eq. (\ref{eq1}) describes the policy gradient for the main environmental task and eq. (\ref{eq2}) describes the policy gradient for the abstract shaping task. We approximate the expectation over the agent's full state ($x_s$) and action ($a_s$) occupancy using rollouts sampled from the agent's experience using its current policy. We use different termination and discount factors $\gamma$ for each gradient computation.
Here, $V_\theta(x_s)$ refers to the value approximation (with parameters $\theta$) at state $x_s$, $v_{s+1}$ of the value target, $r_t$ to the reward, $a_s$ to the motor action, $\pi_{\omega}$ to the policy (with parameters $\omega$), $\gamma$ to the discount factor, and $\rho_s$ to the truncated importance sampling weight, as described by \cite{espeholt2018impala}. $\nabla\mathcal{L}$ is the policy gradient (taken with respect to a proxy loss $\mathcal{L}$ for each task, that corresponds to maximizing the return of that task). All variables in Eq. (\ref{eq1}) with superscript `env' are associated with the environment's motor control task (solving the puzzle or winning the game) while those in Eq. (\ref{eq2}) with superscript `abs' are associated with the auxiliary task in the abstract state space.

In the expert-planner condition, the current abstract state and the target abstract state predicted by the expert planner (as well as the expert's abstract action, in the case of \mujoban) are given as input to the value and policy components of the architecture. The specified abstract states and actions reflect the underlying abstract task. For \mujoban, this observation describes the full state of the \sokoban maze (including the position of the agent, the boxes, and the targets), while the action is the direction of the next movement (north, east, south, or west); the abstract inputs are illustrated in Figure~\ref{fig:mujoban_state}. For \mujogo and \mujoxo, the state describes the full board game, including the position of all pieces. For the board games, the expert does not return an explicit action, as the state is specified in global board coordinates and the action can be inferred directly from the current and next board configuration (this same property is exploited in the design of the agent architectures for board games in \cite{silver2016mastering} and successive work). We use a pre-trained \sokoban RL agent (with the convLSTM-based architecture described in \cite{guez2019investigation}) as the expert planner for \mujoban and we use open-source, off-the-shelf agents for the board games: \gnugo (level 10) \citep{GNUGo2009} for \mujogo and a minimax tree-search agent for \mujoxo. The expert planner is only used to provide target states and actions during agent training: we do not train it or differentiate through it.

The agent receives sensory input from proprioceptive sensors (including touch, position, velocity, and acceleration sensors) as well as cameras (ego-centric and top-down cameras in the case of \mujoban). Each of these inputs are processed via a multi-layer Perceptron (for non-visual sensory inputs) or a residual network \citep{He_2016_CVPR} (for visual inputs) and concatenated before being passed to the recurrent cores. We give details of the observations and architecture in Section \ref{sec:appendix_obs_arch} of the Appendix.

\subsection{Experimental conditions}
\label{sec:conditions}

In the experiments, we probe the effect of different types of expert information on task performance. We modify the architecture as follows in each of the three conditions:

\begin{itemize}{}
    \item \textbf{Expert planner agent:} the agent has access to an expert planner that consistently solves the abstract task embedded in the environment. This agent uses the full architecture described in Section \ref{sec:agent_architecture} and shown in Fig. \ref{fig:mujoban_agent_architecture}, including the reward (and corresponding policy gradient term) reflecting how well the agent follows the expert plan in the abstract space.
    
    \item \textbf{Random planner agent:} the agent receives the current abstract state and a target abstract state chosen from among the possible valid next states. The agent architecture and reward in this case is identical to the previous case; the only difference is that target states are chosen randomly instead of by an expert. Although these random high-level actions sometimes conflict with the goals of the underlying game, we reason that because this extra reward provides information about the underlying game and its dynamics, it may help the agent explore the state space in more structured manner. 

    \item \textbf{Vanilla agent:} the agent has access to minimal prior knowledge about the underlying game and no access to the planner. In the case of \mujoban, the agent is not given the abstract game state. In the case of \mujoxo and \mujogo, the agent is given the current board state to simplify learning, though we expect that similar results could be obtained using only camera inputs. This results in a standard recurrent actor-critic architecture with minimal prior knowledge about the abstract game and no direct information about the dynamics of the game. We train this architecture to maximize the environment reward using only a single value estimate.
\end{itemize}

In \mujoban, we train agents with a curriculum composed of mazes with 5 difficulty levels, cf.\ Appendix Table~\ref{table:mujoban_levels}. During evaluation, we track the agent's performance on each of the difficulty levels separately. In \mujogo, we set the opponent parameters to $\epsilon=0$ and $L=10$ for the GnuGo opponent. In \mujoxo, the opponent plays random moves with probability $\epsilon=0.25$ and plays optimal \tictactoe moves otherwise.

\section{Results}
\label{sec:results}

We report our experimental results below, organized by domain.

\subsection{\mujoban}

\begin{figure}[h]
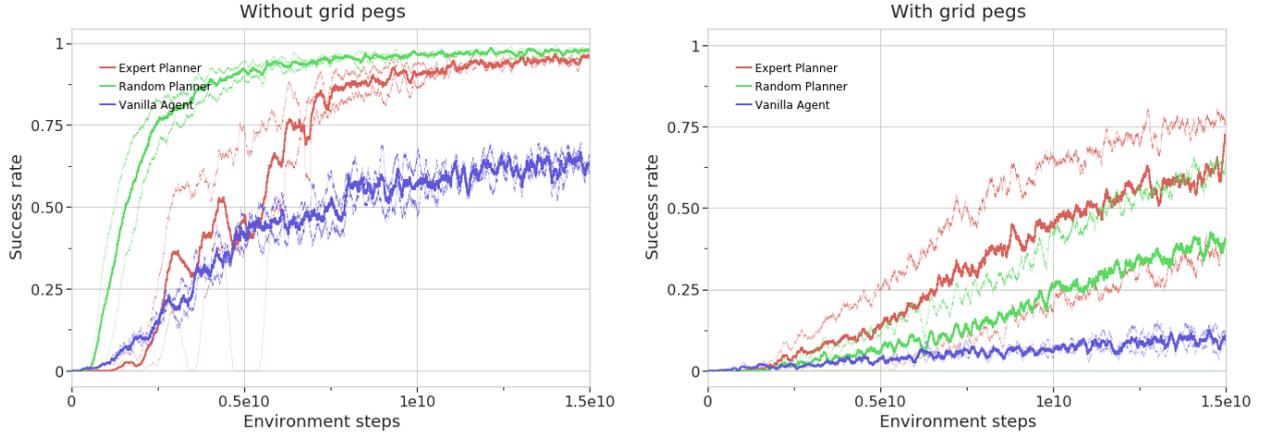

    \begin{subfigure}[b]{0.5\textwidth}
    \centering
      \includegraphics[width=0.98\linewidth]{mujoban_curve_without_pegs.pdf}
      \caption{}
    \end{subfigure}
    \begin{subfigure}[b]{0.5\textwidth}
    \centering
      \includegraphics[width=0.98\linewidth]{mujoban_curve_with_pegs.pdf}
      \caption{}
    \end{subfigure}
	\caption{Agent performance on \mujoban with the larger 10x10 levels (difficulty 5). 
	In the peg-less version (a), the vanilla agent only learned to solve half of the levels after extensive training. The random planner agent performed best in this setting. In (b), where the pegs constrain the agent to solve the task with logical constraints closer to the abstract game, the expert planner performed better and the vanilla planner only ever learned to solve a small fraction of the tasks. Thin traces show the results of each individual seed and thick traces show the mean results across all seeds of each condition. Due to the long wall-clock times associated with running these experiments to convergence, we terminated each run after $1.5\mathrm{e}{10}$ steps (approx. 11 days using 2x2 TPU v2 slices).}
	\label{fig:mujoban_curve}
\end{figure}

Our experimental results for \mujoban confirm that the task is challenging for current RL methods. Learning curves for the larger grids are shown in Figure~\ref{fig:mujoban_curve}, and for other grid sizes in Appendix Figure~\ref{fig:mujoban_smaller_levels}. In the grid peg version, which forces a closer correspondence to the abstract \sokoban problem, agents were unable to solve more than half of the levels, even after extensive training and even though these levels correspond to relatively straightforward abstract problems. Vanilla RL agents, which do not have any abstract augmentation, were inefficient at learning both versions of the task (with and without pegs). This was true even in the presence of a curriculum designed to make it easier for the agent to learn the important skills needed (e.g., how to push boxes to targets). The curriculum provides the agent with a signal that is sufficiently strong to learn the basic mechanics of the task, which suggests that the agent's failure to solve the harder levels is not primarily due to limited exploration. Rather, this suggests that the kind of generalization needed to move beyond greedy behavior (often sufficient to solve simple levels) to longer-term reasoning (needed to solve harder levels) does not emerge in our model-free \mujoban agent. For context, model-free agents can obtain 90\%+ success on the hardest levels in the original, non embodied version of \sokoban \citep{guez2019investigation}. Moreover, the control problem associated with implementing the individual steps of an optimal policy in the abstract task is not particularly difficult (\cite{karkus2020modular} provides some evidence for this assertion). It appears that the RL agent's difficulty in mastering this task arises due to the combination of multi-step reasoning and control, rather than either of these in isolation.

The version without grid pegs was markedly less demanding for all agents, presumably because there is more flexibility to solve the level in an unconventional way (e.g., push boxes diagonally, recover from dead end scenarios). Nevertheless, the model-free agent did not progress much beyond the 50\% mark whereas privileged agents (as described in Section~\ref{sec:Experiments}) achieved good performance (around 90\% of level solved). In the peg-less version, we also found the random planner to perform more robustly than the expert planner --- perhaps because this version aligns less directly with the abstract version of the problem that the expert planner guides towards. In the version with pegs, the agent equipped with the expert planner learned faster -- perhaps because the expert subgoals are often on the optimal solution path in this physics version of the problem. We summarize the final performance on all conditions in Appendix Table~\ref{tab:mujoban_evaluation}.

\subsection{\mujoxo}

\begin{figure}[h]
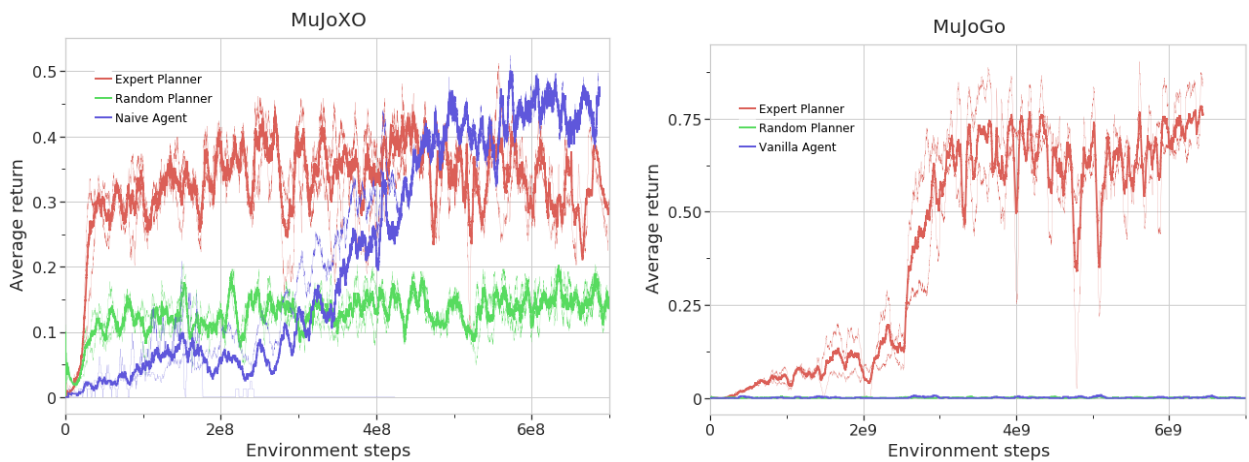

    \begin{subfigure}[b]{0.5\textwidth}
    \centering
      \includegraphics[width=0.98\linewidth]{mujoxo_curve.pdf}
      \caption{}
    \end{subfigure}
    \begin{subfigure}[b]{0.5\textwidth}
    \centering
      \includegraphics[width=0.98\linewidth]{mujogo_curve.pdf}
      \caption{}
    \end{subfigure}
    \caption{Left: Agent performance on \mujoxo. Right: Agent performance on \mujogo (random planner and vanilla agent curves are near 0). In both panels, thin traces show the results of each individual seed and thick traces show the mean results across all seeds of each condition. Please note the difference in $y$ axis scale for each panel.}
    \label{fig:mujo_xo_go}
\end{figure}

\mujoxo has a modest planning load compared to \mujogo and \mujoban, since the abstract problem space of \tictactoe is relatively small (with an average branching factor of 4 and depth less than 9). Nevertheless, careful decision making is still needed given the presence of irreversible decisions and the added requirement of placing moves with a robotic arm. 
Indeed, we observed that our model-free agent required a large number of games (on the order of 1 million) before it could play convincingly, even in this limited planning setting. This is shown in Figure~\ref{fig:mujo_xo_go}-a.
In contrast, the agent equipped with an optimal abstract planner achieved that performance level with around 5 times less games of experience. The agent suggesting random abstract goals, corresponding to placing a legal move at random, was similarly helpful in early training, after which the agent plateaued without progress.
Unlike \mujoban, where physical states have a somewhat fuzzier correspondence to abstract states due to mismatches between the abstract and physical games, the subgoals suggested by the expert in \mujoxo are exactly aligned with the intermediate states of an optimal strategy: the only way to win is to have placed the stone in the right square, and the exact physical position of that stone has then no bearing on the rest of the episode. Our results suggest that knowledge of the abstract structure of the game and its dynamics aids learning, but is not sufficient to encourage the agent to discover how to reliably beat its opponent.

\subsection{\mujogo}

In \mujogo, the additional complexity of planning and the longer game times, considerably increased the learning difficulty for all agents. The vanilla (and random planner) agents did not show any sign of progress after billions of steps of training. Since the agent plays against a fixed-strength opponent in \mujogo, exploration is a notable issue early on in training: the agent needs to observe wins in order to correctly adapt its actions. For the vanilla model-free agent, wins happened very infrequently (approximately 1 in every 1000 games), which might be sufficient to explain the poor performance we observed.
The agent equipped with an abstract planner fared considerably better, eventually beating the fixed opponent in this 7x7 \go game, but it is still remarkably inefficient at learning, considering the small size of the board and the limited strength of its opponent. It required around 4M games to reach a win rate of $\sim$ 60\% (the level of performance at the end of training reported in Figure~\ref{fig:mujo_xo_go}). In comparison, 4M games is approximately the number of self-play games used by \textsf{AlphaGo Zero} to reach professional level on 19x19 boards in the abstract version of the problem \citep{silver2017mastering}.

\section{Conclusion}

Advances in deep learning have spurred profound progress in the field of reinforcement learning. The use of powerful function approximators in the form of large and deep neural networks has allowed machine learning practitioners to move away from hand-crafted features and to solve RL problems that require complex generalization for perception and control. However, many of these successes are limited to problems where decisions and control take place over relatively short time scales, and actions are taken in an abstracted state and action space tailored to the task. Problems that require reasoning and decision making over long time scales using sensoriomotor control cannot yet be solved in an end-to-end fashion. These problems arise frequently in human behavior but are still hard to frame and rarely studied in a controlled experimental setting.

We believe that board games, which are among the most widely used test beds for studying planning capabilities in AI \citep{yannakakis2018artificial}, are also well-suited for study in the embodied control context. While games like these are usually studied in an abstract form, they typically originate from real-world games that employ physical objects with real pieces played using human actuation. These physical version of a board game naturally capture the challenges of coupling abstract reasoning with motor control. They simultaneously provide well defined problems that are known to benefit from algorithmic advances. Because the abstract structure underlying many board games is well understood, board games allow the researcher to easily tune the difficulty of the reasoning task and to analyze the strategies learned by agents.

To encourage research in this direction, we have introduced a set of board games (\tictactoe, \go, and \sokoban) embedded in physical simulations. We have demonstrated that a standard deep RL algorithm struggles to solve these games when they are physically embedded and temporally extended. Agents given access to explicit expert information about the underlying state and action spaces can learn to play these games successfully, albeit after extensive training. We present this gap as a challenge to RL researchers: What new technique or combination of techniques are missing from the standard RL toolbox that will close this gap? We open source these environments to spur development of agents that can reach high level of performance, without privileged information.

Humans appear to use a variety of approaches to help them reason about and solve physically embedded and temporally extended planning problems. They often receive hints about important sub-goals, directly from others in the form of demonstrations, or through culturally-specified cues such as the reuse of conventions in game boards, or abstract rule specifications via language. Humans also appear to discover potentially interesting structure through intrinsically motivated exploration and curiosity. By exploring and exploiting this structure, humans are able to observe, reason, and control in a more directed fashion. We hope that the environments provided here will spur research into how to coherently introduce these capacities into the next generation of RL agents.

\section{Acknowledgments}

We thank Yuval Tassa and Animesh Garg for helpful feedback on earlier versions of the paper.

\bibliographystyle{abbrvnat}
\setlength{\bibsep}{5pt} 
\setlength{\bibhang}{0pt}
\bibliography{bibref_definitions_long,main}

\newpage
\appendix

\section{Code and Environment details}
\label{sec:appendix_code}

The code for all environments is available at \href{https://github.com/deepmind/dm\_control}{https://github.com/deepmind/dm\_control}. Below we provide examples and details for each environment. Each code example relies on the libraries included in the \textsf{dm\_control} package \citep{tassa2020dmcontrol}, including \textsf{PyMJCF} and \textsf{Composer}. 

\subsection{\mujoban}

The code snippet below shows how to instantiate a \mujoban environment. \mujoban is a \textsf{composer} task. As arguments it requires a walker and maze. Here, the maze is constructed using one of the previously released \textsf{BoxoBan} levels (\href{https://github.com/deepmind/boxoban-levels}{https://github.com/deepmind/boxoban-levels}).
And the walker provided is jumping ball walker from dm\textunderscore control.locomotion.

\begin{lstlisting}[language=Python]
from dm_control import composer
from dm_control.locomotion import walkers
from physics_planning_games.mujoban.mujoban import Mujoban
from physics_planning_games.mujoban.mujoban_level import MujobanLevel
from physics_planning_games.mujoban.boxoban import boxoban_level_generator

walker = walkers.JumpingBallWithHead(add_ears=True, camera_height=0.25)
maze = MujobanLevel(boxoban_level_generator)
task = Mujoban(walker=walker,
               maze=maze,
               control_timestep=0.1,
               top_camera_height=96,
               top_camera_width=96)
env = composer.Environment(time_limit=1000, task=task)
\end{lstlisting}

\noindent \mujoban makes use of the \textsf{LabMaze} utility \citep{labmaze, beattie2016dmlab} to create textured 3D mazes from simple ASCII descriptions. Wall and floor textures of different colors are sampled randomly at the start of every episode.

\subsection{\mujoxo and \mujogo} 

\begin{lstlisting}[language=Python, upquote=true]
from physics_planning_games import board_games

environment_name = 'go_7x7'
env = board_games.load(environment_name=environment_name)
\end{lstlisting}

\noindent The simulated \jaco arm \citep{campeau2017jaco} used within these environments is part of the dm\_control suite \citep{tassa2020dmcontrol}.

\section{Experimental details}
\label{sec:appendix_obs_arch}

\newcommand{\resblock}[2]{\(\left[\begin{array}{c}\text{ReLU}\\ \text{conv(3$\times$3, #1)}\\ \text{ReLU}\\ \text{conv(3$\times$3, #1)} \end{array}\right]\)$\times$#2}

\begin{table}[!ht]
    \centering
    \caption{\mujoban difficulty level settings. All levels smaller than $10\mytimes10$ are padded with walls.}
    \begin{tabular}{c|c|c|c}
         Level difficulty & Grid size & Number of boxes & Training ratio \\
        \hline
        \hline
        1 & $5\mytimes5$ & 1 & 0.25 \\
        2 & $7\mytimes7$ & 1 & 0.25 \\
        3 & $7\mytimes7$ & 2 & 0.2 \\
        4 & $8\mytimes8$ & 3 & 0.2 \\
        5 & $10\mytimes10$ & 4 & 0.1\\
    \end{tabular}
    \label{table:mujoban_levels}
\end{table}

\subsection{State estimation in \mujoban}

We estimate the abstract state from the physical scene in order to provide the symbolic input to the expert planner. 
We use the true physical location of the objects and round them off to their nearest grid positions. To prevent overlaps, we first estimate the boxes' positions and then the player location. If an overlap occurs (around grid boundaries), we keep the old player's abstract position.

\subsection{Expert Planner in \mujoban}
The agent has a time limit of $T^{abs}$ to reach the given target state. When the agent reaches the target state or $T^{abs}$ has been reached, we do another state estimation and a new target is proposed by expert planner. We keep the abstract state and target state constant within the period of length $T^{abs}$. The value we found to work best was $T^{abs}=50$.

\subsection{Learning setup}

We train in a distributed RL setup using 1000 actors generating trajectories chunked in unrolls of length 100. Parameters are updated in the learner using the RMSProp optimizer accumulated over batch sizes 64 (\mujoxo and \mujogo) or 128 (\mujoban).

\subsection{Other experimental details}
\label{sec:other_experimental_details}
Task time limits are 900 control steps for \mujoban and \mujogo, and 600 for \mujoxo. Each control timestep is 0.05 seconds for all domains.
The discount factors used for all domains and auxiliary tasks are listed in Table~\ref{table:discounts}.

\begin{wraptable}{r}{5cm}
    \caption{Discount parameters for each domain used in the experiment of Section~\ref{sec:Experiments}.}
    \begin{tabular}{c|c|c}
         Domain & $\gamma^{env}$ & $\gamma^{abs}$ \\
        \hline
        \hline
        \mujoban & 0.99 & 0.9   \\
        \mujoxo & 0.99  & 0.98  \\
        \mujogo & 0.99  & 0.98  \\
    \end{tabular}
    \label{table:discounts}
\end{wraptable}

All cameras (top-down, egocentric, or front-facing) provide RGB inputs with resolution 96x96 to the agents.
For \mujoxo and \mujogo, the proprioceptive inputs are the arm and hand joints' torque, position and velocity, as well of the position of the pinch site. For \mujoban, the proprioceptive inputs are the joints' torque, velocity and position, as well as a touch sensor, accelerometer, gyroscope, and velocimeter.

In all experiments, we used network architectures that have been successfully used for other continuous control tasks using vision \citep{merel2018hierarchical, merel2020reusable}. \textbf{Sensory encoders}: we process all input images using a simplified residual network \citep{He_2016_CVPR}, whose parameters are given in Table \ref{tab:resnet_arch}. All proprioceptive inputs are concatenated, passed through an \textsf{arcsinh} nonlinearity to mitigate the effect of large magnitude values, and then processed by a single linear layer with a ReLU activation. When used as an input, the planner state is processed by a single convolutional layer with a 4$\times$4 kernel and 100 units and a ReLU activation function. The output of this layer is flattened and passed through a linear layer with 256 units and a ReLU activation function. The output of all sensory encoders, as well as the state and expert planner output, are concatenated before being passed to the value and policy cores. \textbf{Value and policy architecture}: the value core is a long-short term memory (LSTM) recurrent neural network \citep{hochreiter1997lstm} with 256 units and the policy core is an LSTM with 128 units. The output of the policy core parameterizes the mean and variance of a Gaussian distribution. Before being used to parameterize the Gaussian, the mean produced by the network is passed through a hyperbolic tangent nonlinearity, and the variance produced by the network is passed through a sigmoid nonlinearity and scaled to $[0.1, 1]$.

We tuned the scale of the expert reward, the learning rate, and the two discount factors separately for each task using ad hoc grid sweeps. The experiments reported in here were run using the best combination of parameters we discovered using at least two random seeds (for \mujoban, \mujoxo, and \mujogo).

\clearpage
\section{Additional figures}

\begin{table}[h!]
\centering
\caption{The residual network architecture used to process images in all experiments. The image resolution is reduced by half at each residual block group. Residual connections are made by adding the input of each residual block (in square brackets) to its output. Unlike other resnets, this architecture does not include any normalization layers.}
    \begin{tabular}{|c|c|}
        \hline
        input size & layer group \\
        \hline
        & conv(3$\times$3, 16) \\
        96$\times$96  & max\_pool(3$\times$3, stride 2) \\
        & \resblock{16}{2} \\
        \hline
        & conv(3$\times$3, 32) \\
        48$\times$48  & max\_pool(3$\times$3, stride 2) \\
        & \resblock{32}{2} \\
        \hline
        & conv(3$\times$3, 32) \\
        24$\times$24  & max\_pool(3$\times$3, stride 2) \\
        & \resblock{32}{2} \\
        \hline
        12$\times$12  & ReLU \\
        \hline
    \end{tabular}
\label{tab:resnet_arch}
\end{table}

\begin{figure}[h!]
  \centering
  \includegraphics[width=0.8\linewidth]{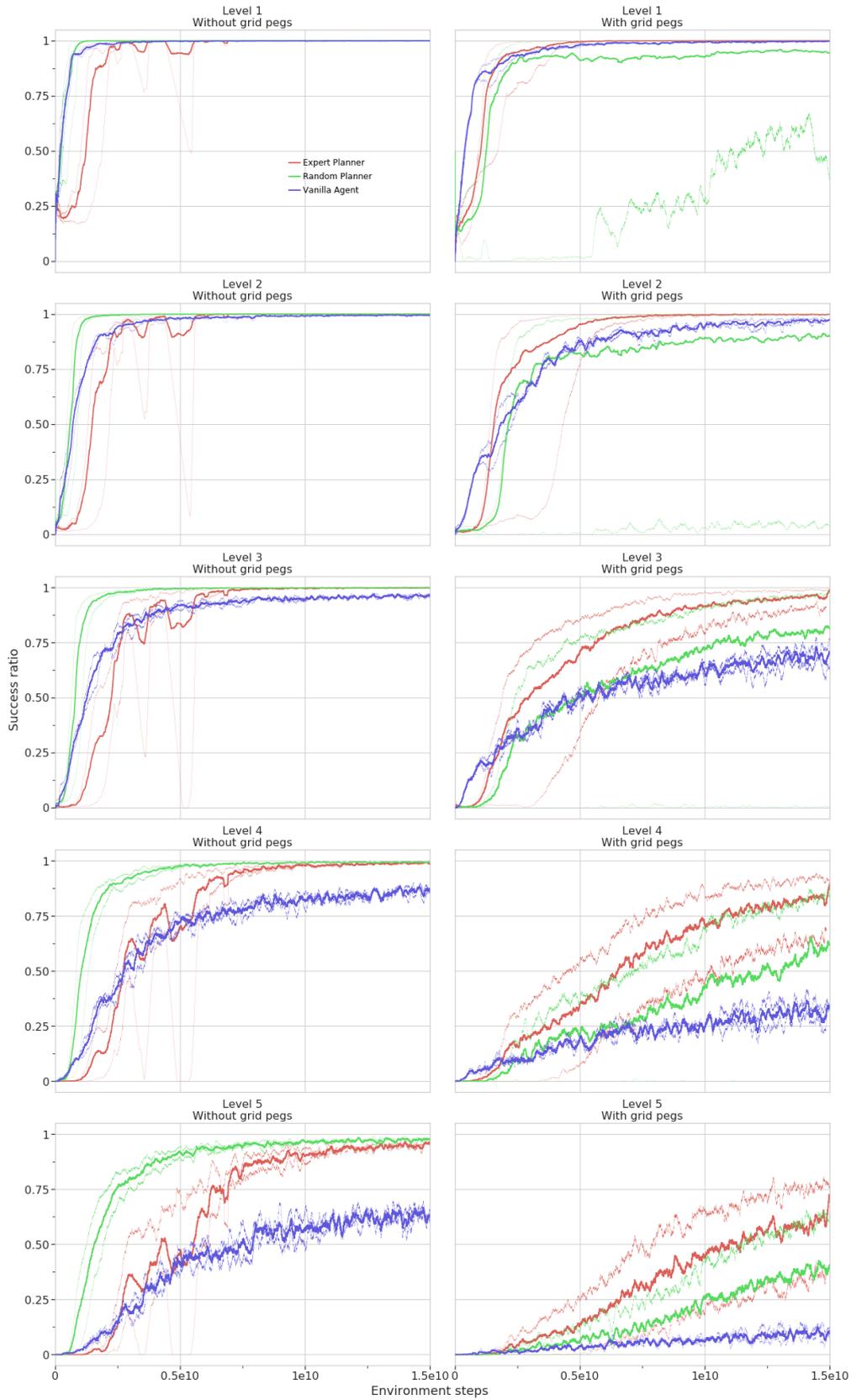}
  \caption{Learning curves for all \mujoban difficulty levels.} 
  \label{fig:mujoban_smaller_levels}
\end{figure}

\begin{table}[h!]
\centering
\caption{Final performance of agents in different settings of \mujoban. The values reflect success ratios over the last 5000 episodes at the end of $1.5\mathrm{e}{10}$ training steps, averaged over agents trained with different seeds.} 
\begin{tabular}{ll|ccccc}
\toprule
\textbf{Setting} & \textbf{Agent} & \textbf{Level 1} & \textbf{Level 2} &  \textbf{Level 3} & \textbf{Level 4} & \textbf{Level 5} \\
\midrule
Without grid pegs   &  Expert Planner  & 100.0\%    & 100.0\%  & 99.9\%  & 99.0\%  & 95.0\%  \\
Without grid pegs   &  Random Planner  & 100.0\%    & 100.0\%  & 99.9\%  &  99.5\% &  97.7\% \\
Without grid pegs   &  Vanilla Agent  & 100.0\%   &  99.5\% &  96.2\% & 86.2\%  &  62.9\% \\
\midrule
With grid pegs   &  Expert Planner  & 100.0\%   &  99.8\% &  94.8\% &  78.2\% &  54.6\% \\
With grid pegs   &  Random Planner  & 75.2\%   & 52.0\%  & 48.7\%  &  42.0\% &  30.2\% \\
With grid pegs   &  Vanilla Agent  &  99.8\%  &  97.2\% & 68.4\%  &  31.5\% & 9.4\%  \\
\bottomrule
\end{tabular}
\label{tab:mujoban_evaluation}
\end{table}

\end{document}